\documentclass[letterpaper, 10 pt, conference]{ieeeconf}  

\IEEEoverridecommandlockouts                              

\usepackage{authblk}
\usepackage{amsmath}
\usepackage{amssymb}
\usepackage{amsfonts}
\usepackage{amstext}
\usepackage{algorithmic}
\usepackage{graphicx}
\usepackage{textcomp}
\usepackage{todonotes}
\usepackage{xcolor}
\usepackage{xspace}
\usepackage{subfigure}
\usepackage{subfloat}
\usepackage{multirow}
\usepackage{wrapfig}
\usepackage{graphicx}
\usepackage{cleveref}
\usepackage{tikz}
\usepackage{url}
\usepackage{flushend}

\newcommand{\name}{FlexLoc\xspace}

\title{\bf \textbf{\name: Conditional Neural Networks for Zero-Shot Sensor Perspective Invariance in  Object Localization with Distributed Multimodal Sensors}}

\author[1]{Jason Wu$^*$\thanks{*The first two authors contributed equally to this paper. Contact: \text{jaysunwu@g.ucla.edu, wangzq312@g.ucla.edu, mbs@ucla.edu}}}
\author[1]{Ziqi Wang$^*$}
\author[1]{Xiaomin Ouyang}
\author[1]{Ho Lyun Jeong}
\author[2]{Colin Samplawski}
\author[3]{\\Lance M. Kaplan}
\author[2]{Benjamin Marlin}
\author[1,4]{Mani Srivastava$^\dag$\thanks{$\dag $Mani Srivastava holds concurrent appointments as a Professor of ECE and CS (joint) at the University of California, Los Angeles and as an Amazon Scholar. This paper describes work performed at the University of California, Los Angeles and is not associated with Amazon.}}
\affil[1]{University of California, Los Angeles}
\affil[2]{University of Massachusetts Amherst}
\affil[3]{US DEVCOM Army Research Laboratory}
\affil[4]{Amazon}

\begin{document}

\maketitle
\pagenumbering{arabic} 
\thispagestyle{plain}
\pagestyle{plain}

\begin{abstract}

Localization is a critical technology for various applications ranging from navigation and surveillance to assisted living. 
Localization systems typically fuse information from sensors viewing the scene from different perspectives to estimate the target location while also employing multiple modalities for enhanced robustness and accuracy. Recently, such systems have employed end-to-end deep neural models trained on large datasets due to their superior performance and ability to handle data from diverse sensor modalities. However, such neural models are often trained on data collected from a particular set of sensor poses (i.e., locations and orientations). During real-world deployments, slight deviations from these sensor poses can result in extreme inaccuracies. 
To address this challenge, we introduce \name, which employs \emph{conditional neural networks} to inject node perspective information to adapt the localization pipeline. Specifically, a small subset of model weights are derived from node poses at run time, enabling accurate generalization to unseen perspectives with minimal additional overhead.
Our evaluations on a multimodal, multiview indoor tracking dataset showcase that \name improves the localization accuracy by almost 50\% in the zero-shot case (no calibration data available) compared to the baselines. The source code of \name is available in \url{https://github.com/nesl/FlexLoc}.

\end{abstract}


\section{Introduction}
\label{sec:intro}


Modern robotic and automation applications are heavily dependent on the ability to localize and track a target with a high degree of accuracy. 
For example, an automated warehouse relies on accurate localization of robots, human workers, and packages to maximize productivity with intelligent scheduling~\cite{rohrig2008tracking}.
A delivery or cleaning robot within a smart building can also rely on the building's localization infrastructure to guide it through a complex floor map~\cite{lin2016enhanced}. 
Finally, various healthcare applications are also greatly interested in the location of a human subject, deriving behavioral biomarkers from subject location to provide early warnings for chronic diseases \cite{ouyang2023admarker}. 

With these numerous groundbreaking applications all depending heavily on accurate localization, recent works have explored the utilization of both \emph{localization infrastructure} and \emph{egocentric} localization technologies. 
Localization infrastructure integrates sensors into the environment, contrasting with egocentric localization that relies upon an on-board suite of sensors to perform techniques such as SLAM~\cite{whyte2006SLAM}. Egocentric localization techniques have been well-studied and achieve exemplary performance~\cite{chen2018SLAM}. Nevertheless, egocentric sensing is susceptible to occlusions and SWaP constraints (size, weight, and power) that can limit the ability of robots to carry sensors. In this paper, we focus on infrastructure-based localization, which does not impose requirements upon the targets and enables ubiquitous localization for non-cooperative human and robotic targets.

In this work, we employ a set of distributed (multi-view) multimodal sensors for object localization. 
Existing localization infrastructures typically incorporate multiple sensor nodes with partially overlapping sensor perspectives as a \textbf{multi-view} system to improve localization accuracy and expand sensing coverage. These systems estimate the target's location in a global coordinate system using camera~\cite{samplawski2023heteroskedastic}, UWB radar~\cite{bocus2021passive}, or acoustic sensors~\cite{de2010localization}. Moreover, in real-world scenarios, the perception of a target is heavily influenced by environmental factors, and a single sensor modality is often inadequate in capturing sufficient information to enable robustness. For example, RGB cameras fail to provide relevant information in low-lighting conditions. Therefore, several new \textbf{multimodal} sensing systems have been proposed to enhance the reliability of indoor localization by fusing information across multiple distinct sensors~\cite{liu2017multiview, zhao2019enhancing,amri2015indoor}. For instance, Zhao et al. \cite{zhao2019enhancing} enhances camera-based indoor localization with inertial measurement units (IMUs) and WiFi CSI signals.\looseness=-1



To handle multi-view multimodal sensor data, localization systems use large-scale end-to-end deep learning models to predict target locations as these data-driven models can handle various sensor modalities and yield superior performance~\cite{vakil2021survey}. These models can be divided into \textbf{early-fusion} (feature-level fusion) and \textbf{late-fusion} (result-level fusion) approaches~\cite{gadzicki2020early}. Early fusion approaches extract intermediate features from each sensor and fuse them together to generate one prediction. Meanwhile, late fusion approaches predict the target location individually using each sensor, generating a prediction by reconciling all the results (e.g., using a Kalman Filter). While late-fusion approaches can minimize the communication overhead in a distributed sensing system, early-fusion approaches often achieve superior performance, as the model facilitates the early collaboration of sensors with different strengths.

\begin{figure}[t!]
    \centering
    \includegraphics[width=\linewidth]{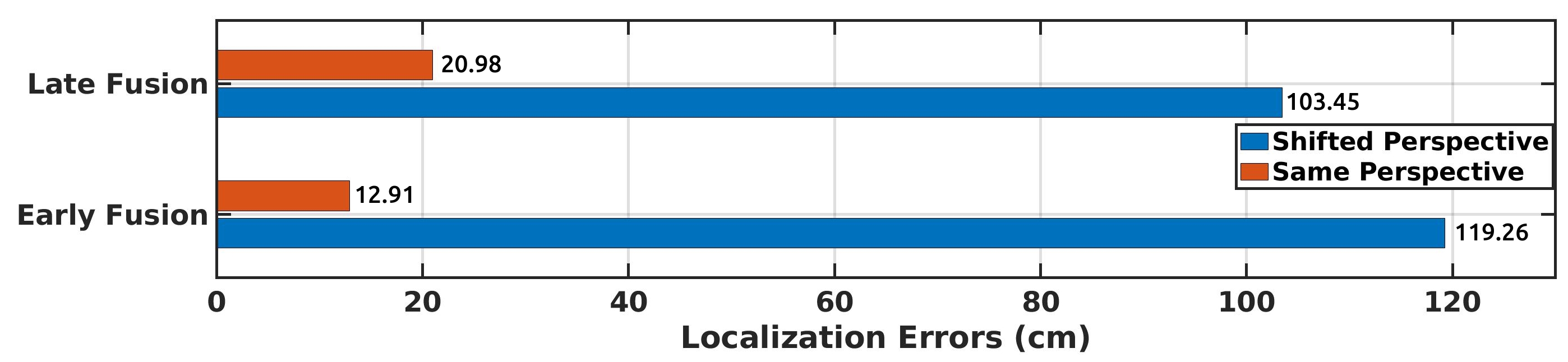}
    \vspace{-23pt}
    \caption{Performance degradation due to sensor perspective shift.}
    \label{fig:motive}
    \vspace{-20pt}
\end{figure}

\textbf{Problem and Motivation:} Despite the increased robustness towards environmental factors provided by learning-enabled multimodal systems, current approaches fail to address the critical issue of \textbf{sensor perspective shift}. Deep learning localization models are often trained on datasets collected using a fixed set of sensor positions and orientations (i.e., sensor perspectives), and thus the correspondence between sensor data and the target global coordinates is inherently embedded in the neural network weights. This correspondence breaks down when the model is applied to data collected from an unseen configuration of sensor placements. We illustrate this problem here through a motivational study. 
We leverage GDTM~\cite{wang2023gdtm}, a multi-hour dataset we collected using distributed multimodal sensor nodes to track a remote controlled car in indoor environments. In this dataset, we introduce diversity in sensor perspectives by altering the position and orientation of sensor nodes between trials. We train both an early-fusion and a late-fusion model on data from one sensor perspective, with the results shown in Figure~\ref{fig:motive}.
\noindent When we evaluate both models on new data from the \emph{same} viewpoint, we achieve reasonable localization performance. However, both models fail spectacularly when evaluated on sensor data collected from a new sensor placement, with average distance errors of $>$1~\!m.\looseness=-1

This challenge of sensor perspective shift poses a significant barrier to the practicality of localization systems. Not only can sensor pose gradually shift over time, factors such as relocation after maintenance are inevitable during the deployments. Attempting to train a new system from scratch for each new set of sensor perspectives is highly impractical due to the large amount of data required. Thus, we must \textbf{design models that can automatically adapt to sensor perspective shift without additional calibration data} (i.e., in a zero-shot manner). GDTM~\cite{wang2023gdtm} attempted a solution with more than 30~\!cm of error on new perspectives with a suboptimal late-fusion approach: localizing the target at each distributed sensor within its local coordinate system and applying a hard-coded coordinate transform into global coordinates.\looseness=-1

\begin{figure}
    \centering
    \includegraphics[width=0.8\linewidth]{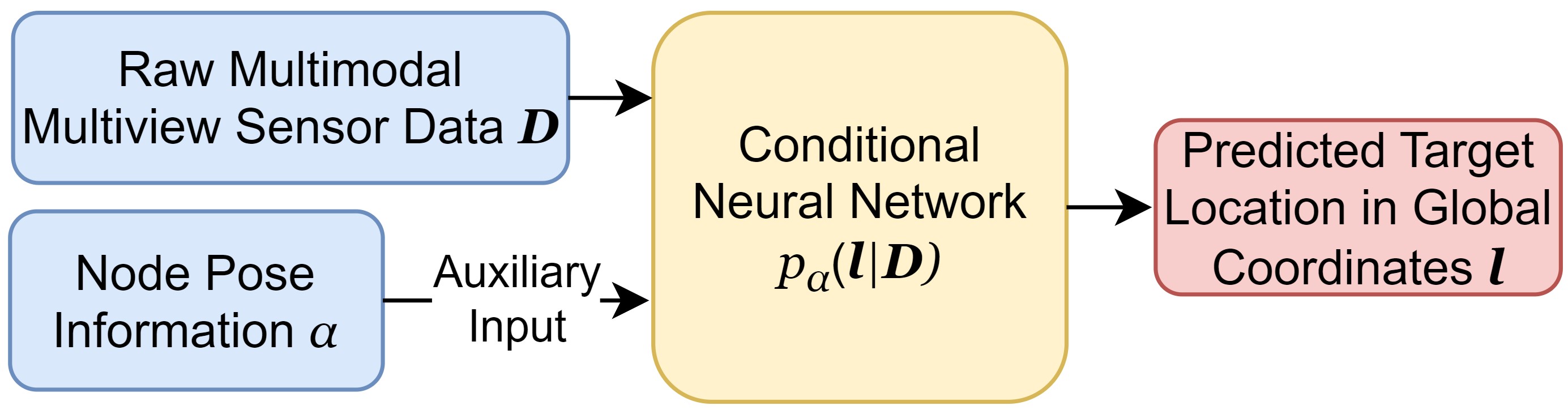}
    \vspace{-10pt}
    \caption{Illustration of our key idea to enable perspective invariant object localization. Unlike existing approaches that are unaware of sensor perspectives, we inject node pose information into the network through conditional neural networks.}
     \vspace{-15pt}
    \label{fig:basic-idea}
\end{figure}

\textbf{Proposed Method:} To address this challenge, we propose \name, a new multimodal multi-view sensor fusion approach that is robust to the perspective shift of nodes within a distributed multimodal sensor network, and is compatible with the end-to-end early fusion approach. As shown in Figure \ref{fig:basic-idea}, our main idea involves injecting node orientation and position (pose) information into the localization network. During the training process, the model is exposed to a large variety of viewpoints with associated pose information to learn how the sensor's pose affects the prediction of global coordinate location from multimodal sensor data. Then, during test time, the network utilizes the supplied pose information to make accurate predictions from unseen sensor viewpoints.\looseness=-1 


Specifically, \name builds upon a feature-level (early) fusion architecture primarily consisting of backbones, adapters, a transformer encoder, and an output head. From a high level, the backbones (implemented using ResNets and ViTs~\cite{dosovitskiy2020vit}) extract features from raw sensor data. The features are mapped to fixed-sized vectors by the adapters and then aggregated. A multi-layer transformer encoder then processes the aggregated features. Finally, the output head yields the predicted global coordinates. To inject the node pose information, we design and insert \textbf{Conditional Neural Networks} into the aforementioned architecture. The key idea is to augment the main neural network with a small subset of additional weights conditioned on auxiliary inputs (sensor pose in our scenario). These dynamic weights are generated from the auxiliary input using a lightweight neural network during both training and runtime. After model training, the conditional neural networks can leverage supplementary auxiliary information during inference at runtime, thereby achieving greater robustness over the unconditional models relying solely on the original sensor data input. \looseness=-1

Conditional neural networks have elevated the performance of applications such as Visual Question Answering or Speech Enhancement~\cite{vries2017CBN, zhang2021CLN, strub2018FILM}. With \name, we seek to transfer this idea to the sensor perspective shift problem, which contains increased complexity due to the fragility of the learned 3D spatial relationships. Specifically, we create two variants of conditional neural networks. First, drawing inspiration from the multiplicative nature of coordinate transforms, we propose a \emph{Conditional 1D Convolution} (CondConv) whose convolutional weights are conditioned on the sensor pose. These CondConv layers are inserted after the adapter to transform the extracted sensor features with the sensor pose information. Second, we propose a lightweight design for injecting the pose information through \emph{Conditional Layer Normalization} (CLN), given the pervasive existence of normalization layers in the backbone models. We replace the learnable parameters in the normalization layers, i.e., scale ($\gamma$) and offset ($\beta$), with values derived from the pose. The lightweight CLN approach is the better choice when computing resources are limited, while the CondConv method performs better in various complex scenarios.

To evaluate \name, we leverage our previously collected multi-modal multi-view dataset for indoor geospatial vehicle tracking~\cite{wang2023gdtm}. Our dataset contains a total of nine-hours of data collected from three nodes that are equipped with four sensor modalities (RGB camera, depth camera, mmWave radar, and microphone), with 22 groups of unique sensor node perspectives (one perspective is an arrangement of 3 nodes). Our results in Section~\ref{sec:experiments} show that our approach can achieve good zero-shot performance on data collected in unseen perspectives while also exhibiting test-time robustness to different combinations of sensor modalities.


In summary, we make the following key contributions:
\begin{itemize}
    \item We propose two methods for improving test-time performance of localization models on unseen sensor perspectives: Conditional 1D Convolution and a lightweight design based on Conditional Layer Normalization. This work constitutes the first exploration into applying conditional neural networks to address test-time robustness in localization models. 
    \item We independently integrate the proposed two conditional methods into an early fusion architecture for multimodal multi-view learning. We then perform a detailed study investigating the performance of the two methods.\looseness=-1
    \item \name involves a minimal number of additional parameters (at most  0.2\% of model parameters), and its simple implementation allows for easy integration within existing neural networks
    \item \name shows good zero-shot generalization performance to various perspectives while also retaining good performance in the absence of various sensor modalities. 
\end{itemize}


\section{Methodology}
\label{sec:methodology}

\subsection{Multimodal Multiview Dataset}
\label{subsec:dataset}

\begin{figure}
    \setlength{\abovecaptionskip}{-0.15cm}
    \centering
    \includegraphics[width=0.7\linewidth]{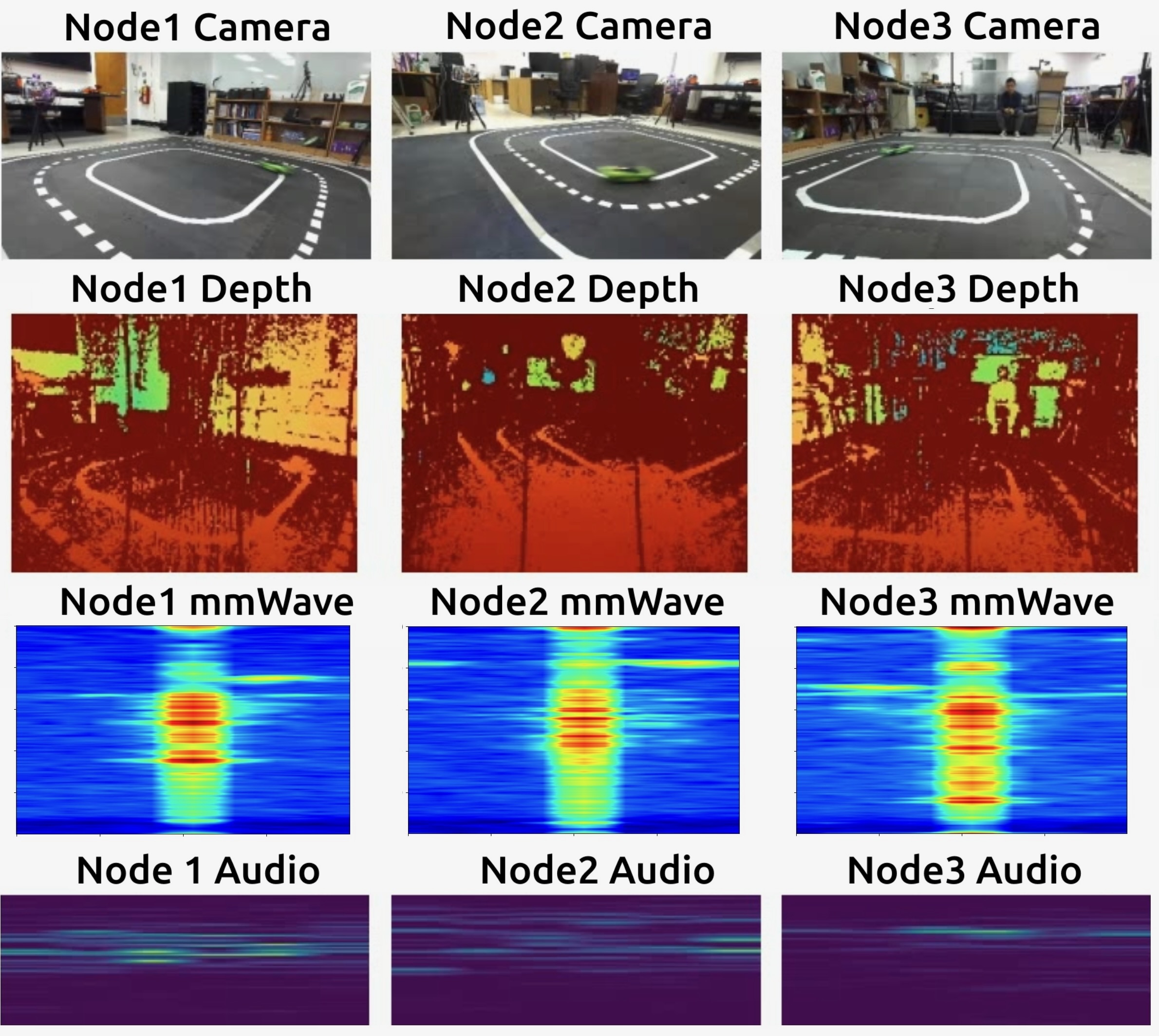}
    \caption{Examples of multi-modal sensor data collected by three nodes.}
    \label{fig:dataset}
    \vspace{-15pt}
\end{figure}

Public multimodal, multiview datasets containing shifting perspectives of sensors are scarcely available. We collected a nine-hour multimodal indoor tracking dataset, namely GDTM~\cite{wang2023gdtm}, to facilitate the research of sensor perspective shift problem.
This dataset is composed of data from three data collection nodes placed at various locations around an indoor race track to localize a remote-controlled car. Each node is composed of multiple modalities (camera, depth, mmWave, and audio, see Fig.~\ref{fig:dataset} for details). 

The position and orientation of these nodes are adjusted between trials to generate a large diversity of sensor viewpoints (perspectives of the three sensor nodes). We utilize this variety of viewpoints to develop our perspective invariant model. GTDM contains 22 unique viewpoints, allowing us to construct a dataset containing 13 views in the training dataset (105 minutes), 4 views in the validation set (35 minutes, for model development and early stopping of training), and 5 views for testing (55 minutes, for final result reporting). Different modalities and ground truths are time-synchronized at 15~\!Hz. All three sets contain both circular motions and random trajectories of the robot car moving in the scene.

We quantified the similarity between perspectives $V_1$ and $V_2$ by computing the \emph{structural similarity index measure} (SSIM) of the camera data: (1) we take a frame of camera data from each sensor node in $V_1$ and $V_2$, resulting in three images in each view. (2) We compute pair-wise SSIM $\mathbf{S}\in\mathrm{R^{3\times3}}$. (3) We take max($\mathbf{S}$), the SSIM of the most similar image pair, as the similarity between $V_1$ and $V_2$. We introduce a two-stage procedure to ensure that the viewpoints used for testing are different from those for training and validation. First, we made automatic calculations to ensure that the 
similarity score between $V_{train}$, $V_{val}$, and $V_{test}$ are all smaller than 0.60. Second, we print out one frame of camera data from each viewpoint and do a visual comparison to double-check.

\subsection{Overall Model Architecture}
\label{subsec:model_arch}

\begin{figure}
    \setlength{\abovecaptionskip}{-0.1cm}
    \centering
    \includegraphics[width=\linewidth]{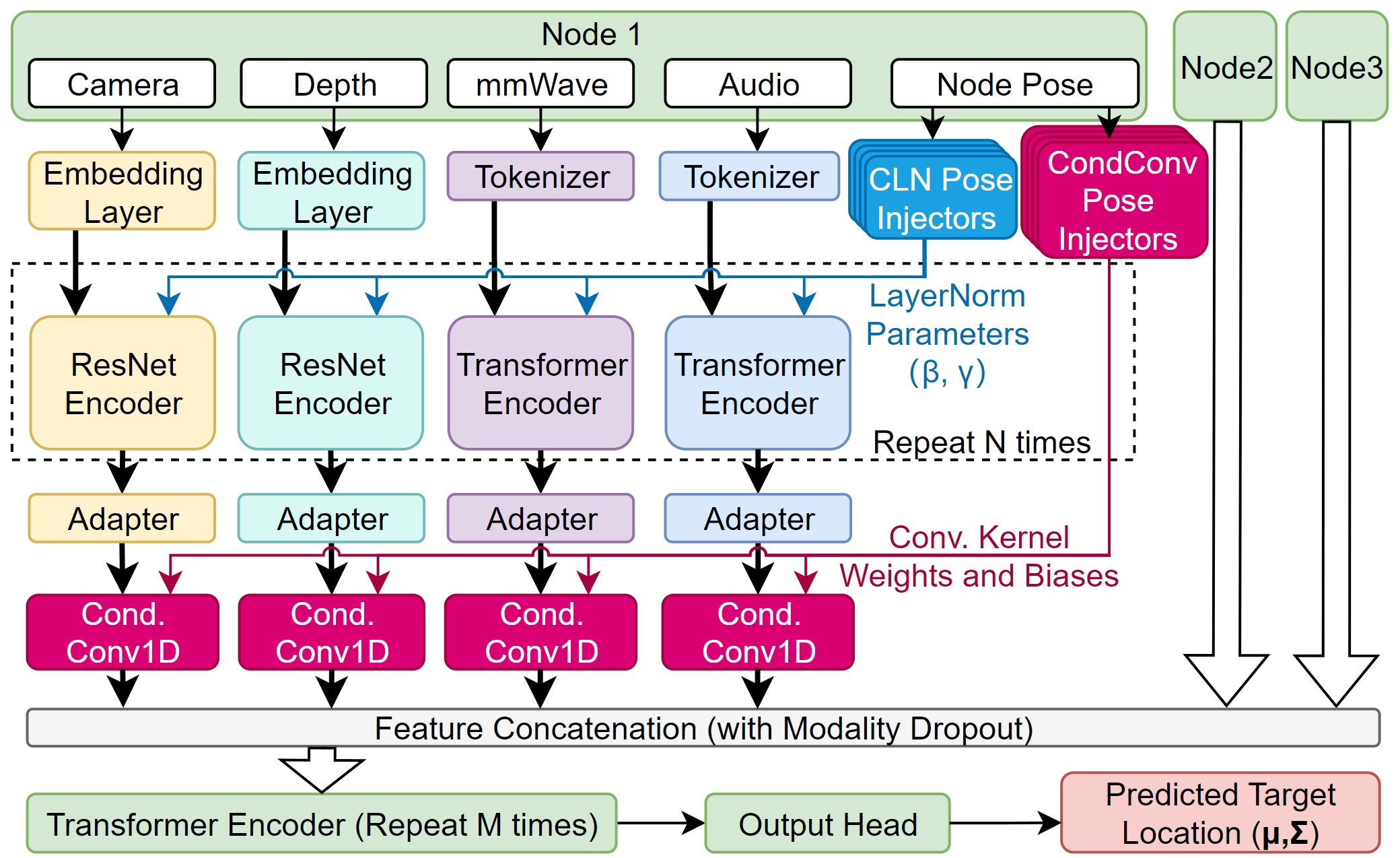}
    \caption{Complete \name architecture containing Conditional Convolution and Conditional Layer Normalization. Conditional Convolution derives its 1D convolutional kernel weights from the sensor pose and transforms the extracted features of the sensor data. Conditional Layer Normalization is a more lightweight design integrated into the backbones, where we replace the learnable parameters $\gamma$ and $\beta$ with values derived from sensor pose.}
    \label{fig:complete_model}
    \vspace{-15pt}
\end{figure}


Figure~\ref{fig:complete_model} showcases the overall architecture of \name, which outputs predictions of the target location in pre-established global coordinates with the help of sensor pose information. First, \name utilizes \emph{backbones} shared across all the nodes to process each modality independently with the goal of extracting key information (embeddings) pertinent to the localization task. The \emph{adapters} (shared among nodes) accept these embeddings from the backbones as input and perform modality-specific processing to transform the distinct feature vectors into vectors of the same dimension. The output of the adapters are then processed by the Conditional Convolution to further introduce additional pose information. Afterwards, the fusion of cross-modal information is performed by a \emph{transformer encoder}. Finally, an \emph{output head} utilizes the information in the joint embedding to make a unified prediction of the car's location in the pre-established global coordinate system. As an alternative to Conditional Convolution, we also propose a lightweight conditional neural network based on Conditional Layer Normalization, which is integrated into backbones to incrementally introduce pose information into the final embeddings of different sensor modalities (drawn in the same figure to save space).

Specifically, our backbones are composed of either Residual Convolutional Networks (ResNet)~\cite{he2016resnet}, or stacks of transformer encoders following the work on \emph{Vision Transformers}~\cite{dosovitskiy2020vit}. ResNet backbones are a good candidate for processing visual data (camera and depth information) due to their ability to extract information about key objects in the scene while ignoring irrelevant background information. In contrast, when processing mmWave and audio sensor data, we employ a simpler stack of transformer encoders due the relative lack of background content. 

Moreover, Conditional Convolution or Conditional Layer Normalization are strategically inserted into this architecture to incorporate pose information. Conditional Convolution operates on the extracted embeddings containing dense information regarding the object's position, injecting the sensor pose information through a series of 1D convolutional kernels. In contrast, Conditional Layer Normalization serves as a lightweight method of incorporating pose information into the early stages of data processing. We replace the traditional Batch Normalization and Layer Normalization components within each backbone with Conditional Layer Normalization that derives its parameters from the sensor pose information. We evaluate the performance of the two conditional methods within Section~\ref{sec:experiments}.\looseness=-1



\subsection{Conditional Convolution}

To effectively inject the sensor pose information, we explore transforming the feature vectors of a particular node (outputs of the adapters in Fig. \ref{fig:complete_model}) with a \emph{conditional 1D convolution}. Given an input vector $\mathbf{x}\in \mathbb{R}^{N}$ from the intermediate features of $M$ sensor modalities $\mathbf{X}=\{\mathbf{x}_0, ..., \mathbf{x}_{M-1}\} \in \mathbb{R}^{M\times N}$
, 1D convolution with $K$ kernels of size $S$ generates an output $\mathbf{y}\in \mathbb{R}^{K\times N}$ according to Equation 1, where $*$ refers to the convolutional operator, 
\begin{equation}
    \mathbf{y}_{k} = b_k + \mathbf{x} * \mathbf{w}_{k} ,
     \quad
    \mathbf{w_k} \in \mathbf{W},
    b_k \in \mathbf{b}.
\end{equation}

\begin{figure}
    \setlength{\abovecaptionskip}{-0.1cm}
    \centering\includegraphics[width=\linewidth]{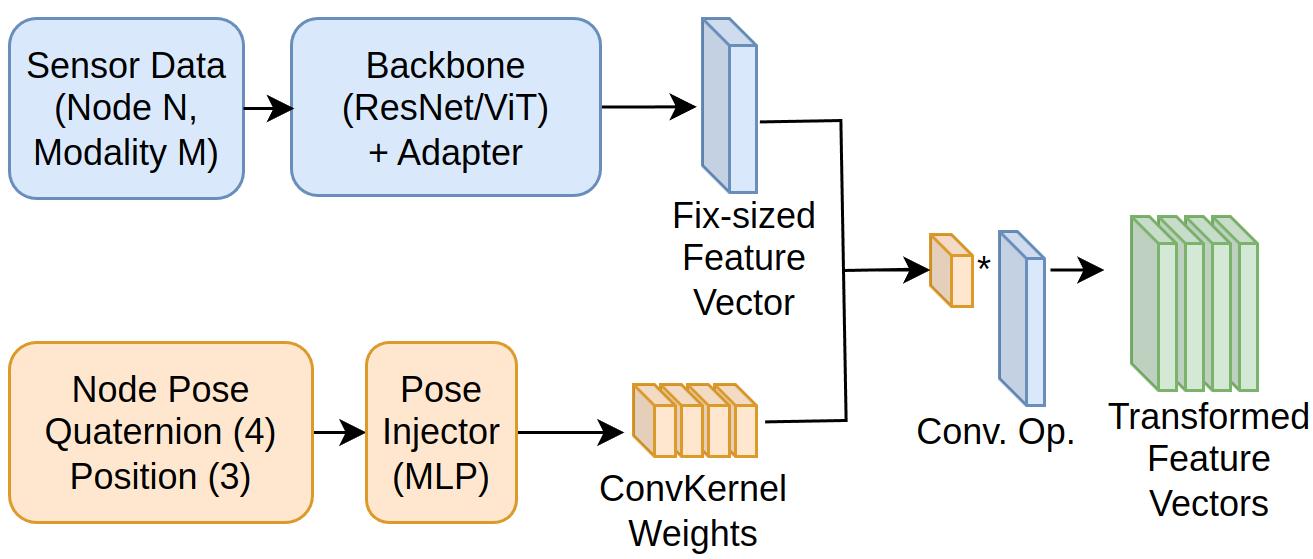}
    \caption{Implementation and Integration of Conditional 1D Convolution.}
    \vspace{-15pt}
    \label{fig:condconv}
\end{figure}

\emph{Conditional} 1D convolution replaces the learnable weights $\mathbf{W}\in \mathbb{R}^{K\times S}$  and biases $\mathbf{b}\in \mathbb{R}^{K}$ with parameters generated from the auxiliary input (sensor pose), fusing the information within $\mathbf{X}$ with additional context. Figure~\ref{fig:condconv} provides a detailed snapshot of how conditional convolution is integrated into the overall system. We convert our node pose into a set of 7 values, including four quaternion values representing rotations and $(x, y, z)$ values representing positions. These values are expanded by a series of linear layers into a higher dimension, from which we extract the weights and biases for the $K$ convolutional kernels of length $S$. These adaptive kernels are convolved with the feature vector to generate $K$ new feature vectors with position information. 

\begin{table*}[]
\centering
\begin{tabular}{|c|cc|cc|cc|cc|cc|cc|}
\hline
                                                                                     & \multicolumn{2}{c|}{View 1}                & \multicolumn{2}{c|}{View 2}                & \multicolumn{2}{c|}{View 3}                & \multicolumn{2}{c|}{View 4}                 & \multicolumn{2}{c|}{View 5}                & \multicolumn{2}{c|}{Average}               \\ \hline
Models Architecture                                                                  & \multicolumn{1}{c|}{Mean}           & Std  & \multicolumn{1}{c|}{Mean}           & Std  & \multicolumn{1}{c|}{Mean}           & Std  & \multicolumn{1}{c|}{Mean}           & Std   & \multicolumn{1}{c|}{Mean}           & Std  & \multicolumn{1}{c|}{Mean}           & Std  \\ \hline
FlexLoc (CondConv)                                                                   & \multicolumn{1}{c|}{\textbf{17.10}} & 0.96 & \multicolumn{1}{c|}{\textbf{14.45}} & 0.68 & \multicolumn{1}{c|}{\textbf{13.30}} & 1.06 & \multicolumn{1}{c|}{\textbf{11.89}} & 1.07  & \multicolumn{1}{c|}{\textbf{14.67}} & 2.03 & \multicolumn{1}{c|}{\textbf{14.28}} & 1.16 \\ \hline
FlexLoc-light (CLN)                                                                        & \multicolumn{1}{c|}{20.50}          & 0.55 & \multicolumn{1}{c|}{24.88}          & 2.78 & \multicolumn{1}{c|}{21.98}          & 4.04 & \multicolumn{1}{c|}{24.38}          & 11.90 & \multicolumn{1}{c|}{18.45}          & 2.33 & \multicolumn{1}{c|}{22.04}          & 4.32 \\ \hline
Baseline-early (Brute Force)                                                           & \multicolumn{1}{c|}{32.61}          & 3.57 & \multicolumn{1}{c|}{24.66}          & 3.27 & \multicolumn{1}{c|}{26.52}          & 3.40 & \multicolumn{1}{c|}{24.25}          & 7.38  & \multicolumn{1}{c|}{26.29}          & 5.78 & \multicolumn{1}{c|}{26.86}          & 4.68 \\ \hline
Baseline-late (Hard-coded Transform) & \multicolumn{1}{c|}{33.69}          & 4.97 & \multicolumn{1}{c|}{25.36}          & 1.16 & \multicolumn{1}{c|}{20.93}          & 0.05 & \multicolumn{1}{c|}{32.97}          & 1.23  & \multicolumn{1}{c|}{24.71}          & 1.94 & \multicolumn{1}{c|}{27.53}          & 1.87 \\ \hline
\end{tabular}
\caption{Euclidean Distance Error (in cm, averaged across 3 seeds), \name vs baselines}
\vspace{-30pt}
\label{table:primary_results}
\end{table*}

\subsection{Conditional Layer Normalization}
\label{subsec:cln}

Layer Normalization~\cite{ba2016ln} is a feature normalization technique that reduces training time and improves generalization of neural networks~\cite{lyu2022generalization}. It performs normalization with the mean and standard deviation computed over elements of the same batch, while also introducing a learnable \emph{per element scale and bias} to improve performance. Equation 2 outlines applying Layer Normalization to a single vector $\mathbf{x}\in \mathbb{R}^{N}$ from a batch of intermediate feature vectors $\mathbf{X}=\{\mathbf{x}_0, ..., \mathbf{x}_{B-1}\} \in \mathbb{R}^{B\times N}$:
\begin{equation}
    \mathbf{y} = \boldsymbol{\gamma} \cdot \frac{\mathbf{x} - \mu}{\sigma} + \boldsymbol{\beta}, \text{  }
    \mu = \frac{1}{N}\sum_{j=0}^{N - 1} x_j, \text{  }
    \sigma = \frac{1}{N}\sqrt{\sum_{j=0}^{N-1}(x_j - \mu)} \\
\end{equation}
with $\boldsymbol{\gamma}\in \mathbb{R}^{N}$ and $\boldsymbol{\beta}\in \mathbb{R}^{N}$ acting as \emph{per element} scale and shift operations.

\begin{figure}
    \setlength{\abovecaptionskip}{-0.1cm}
    \centering
    \includegraphics[width=\linewidth]{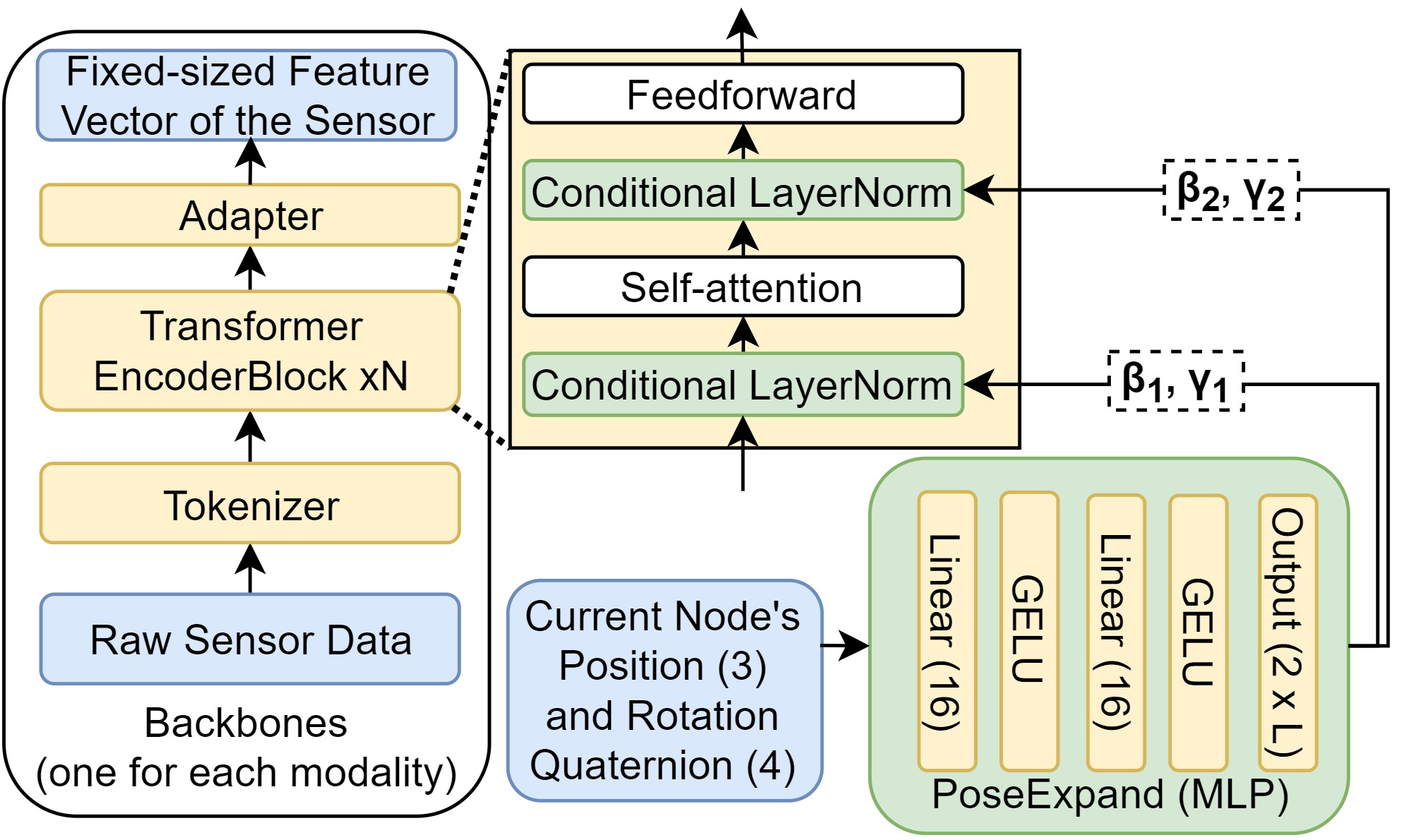}
    \caption{Implementation and Integration of Conditional Layer Normalization.}
    \vspace{-15pt}
    \label{fig:cln}
\end{figure}

We implement \emph{Conditional Layer Normalization} (CLN) by replacing the learnable $\boldsymbol{\gamma}$ and $\boldsymbol{\beta}$ parameters with values derived from the node's pose information. Figure~\ref{fig:cln} illustrates how the layer normalization blocks contained within each transformer encoder of the backbone are replaced with their adaptive counterparts. This same process occurs with the Batch Normalization layers in the ResNet. These adaptive weights are derived from a \emph{separate} set of linear layers (PoseExpand). 
PoseExpand encodes the sensor pose into an vector of dimension 16, from which the unique $\gamma$ and $\beta$ value of each Layer Normalization block are obtained.  Note that to reduce the number of additional parameters, we use \emph{one single} $\gamma$ scalar and \emph{one single} $\beta$ scalar to modify the entire layer instead of performing per element scale and shifts. 
Furthermore, in vanilla LayerNorm, the $\boldsymbol{\gamma}$ and $\boldsymbol{\beta}$ parameters are shared across the batch $B$. In CLN, however, data entries within a batch are not guaranteed to come from the same sensor viewpoint. As a result, we compute one $\gamma_i$ and one $\beta_i$ scalar for each entry $i$ in the batch. Thus, with an input of $\mathbf{X} \in \mathbb{R}^{B\times N}$, every element of $x_{ij}$ from $\mathbf{X}$ will be normalized by $\mu_{i}$ and $\sigma_i$ , then scaled and shifted by $\gamma_i$ and $\beta_i$.

\subsection{Training Strategies}
\label{subsec:pretrain}


\subsubsection{Backbone Pretraining} We utilize a set of pre-trained weights to imbue our backbones with an understanding of how to process the input sensor data. In the camera and depth ResNet-18 backbones, we load a checkpoint pre-trained on ImageNet1K. As we designed the Transformer Encoder backbones ourselves, we chose to pre-train those backbones on 5 minutes of data from a viewpoint in the training dataset. The pre-training process involved separately training each modality's backbone to ensure it has an understanding of how to identify the car from sensor data. 

\subsubsection{Sensor Dropout} Furthermore, we also utilized a \emph{modality dropout} process during the training of the model to fully realize the benefits of the rich multimodal input. We found that training the model with all modalities active resulted in only the most informative modalities contributing to the final result. To combat this, we established a 50\% chance for a given modality to not contribute in a given batch during training, effectively forcing the model to learn with various subsets of modalities. We analyze this in greater detail within Section~\ref{subsec:training_strats}.\looseness=-1

\section{Evaluations}
\label{sec:experiments}

We evaluate a variety of models on the multiview dataset (Section~\ref{subsec:dataset}) to showcase the performance benefit of our conditional networks. We are only concerned with predicting the car's 2-dimensional coordinate, as the car's elevation within the global coordinate system is unchanged. Thus, the primary metric for quantifying the performance is \emph{Average Euclidean Distance} between the predicted location and the ground truth. We evaluated all our models on five \emph{unseen} sensor perspectives.

\subsection{Comparison with Existing Methods}
We test our two conditional models and compare them against two baseline models:
\begin{enumerate}
    \item \textit{Unconditional Early Fusion}: This is essentially \name with no conditional layers.
    \item \textit{Local Coordinate Late Fusion} \cite{wang2023gdtm}: In this approach, each modality and node pair makes an independent prediction of the object location in the node's local coordinate system, after which the predictions are transformed into a global coordinate system via a known transform matrix and merged with a Kalman Filter.\looseness=-1
\end{enumerate}

Table~\ref{table:primary_results} showcases that all the conditional models perform better against the baseline models. In particular, Conditional Convolution (CondConv) achieves the best performance with at least 46\% improvement over the baseline methods. In fact, the localization error of CondConv is comparable to models that are \emph{trained and tested on the same viewpoint}. For example, in Section~\ref{sec:intro}, we trained an early-fusion model and a late-fusion model on a single configuration of sensor viewpoints and tested on unseen data from the same viewpoint, achieving 12.91 cm and 20.98 cm of error, respectively. CondConv achieves an impressive 32\% improvement over the single-perspective late fusion model and comes within 10\% of the single-perspective early fusion model. These results exemplify the robustness of CondConv towards completely unseen sensor perspectives. As this is our best performing model, we term \name as the network containing only CondConv.

While Conditional Layer Normalization (CLN) does not achieve the same level of performance as Conditional Convolution, it can still provide a modest 17.9\% improvement over the baseline early fusion model lacking conditional layers. We hypothesize that its worse performance compared to CondConv is due to the more challenging task of incorporating pose information into the backbone itself. One key advantage of CLN lies in its lightweight nature, where it employs far fewer parameters than CondConv (analyzed in Section~\ref{subsec:params}). Thus, we term CLN as \name-Light. It is also worth noting that \name-Light encounters an unlucky seed where the model performance degrades, which contributed to the high standard deviation.\looseness=-1

\begin{figure}
    \vspace{-1pt}
    \setlength{\abovecaptionskip}{-0.1cm}
    \centering
    \includegraphics[width=0.75\linewidth]{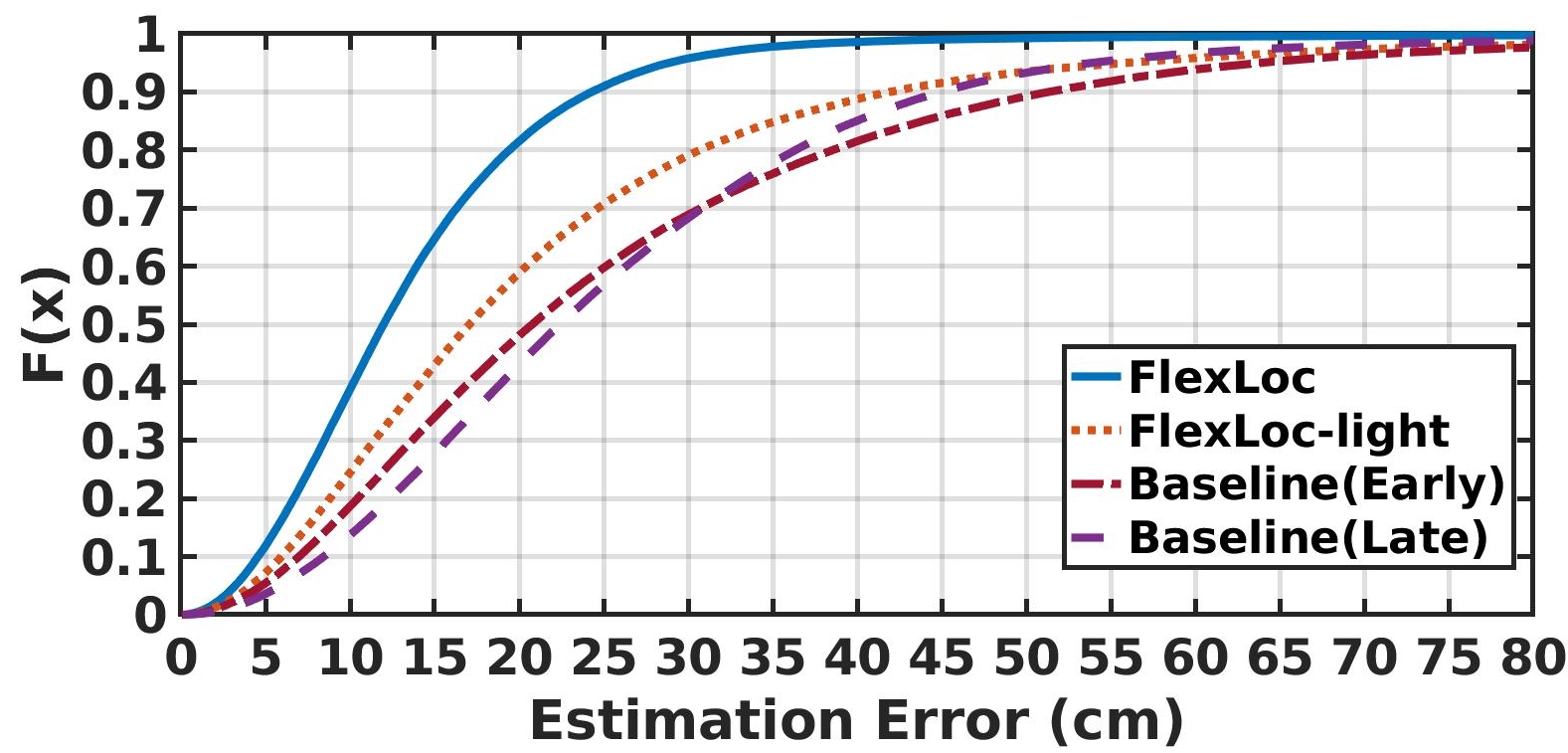}
    \caption{Cumulative distribution function (CDF) plot of frame-by-frame localization error of different methods.}
    \vspace{-15pt}
    \label{fig:cdf}
\end{figure}

Figure~\ref{fig:cdf} showcases a visual depiction of each method's localization error in a CDF. We observe a significant boost in performance between \name and the other methods, while \name-Light maintains a smaller yet notable improvement over the two baseline methods. Another sizable advantage of \name lies in its low 90\% error at only 25 cm, which is 20 cm less error than the next closest baseline method. 

\begin{figure}
    \setlength{\abovecaptionskip}{-0.2cm}
    \centering
    \includegraphics[width=0.98\linewidth]{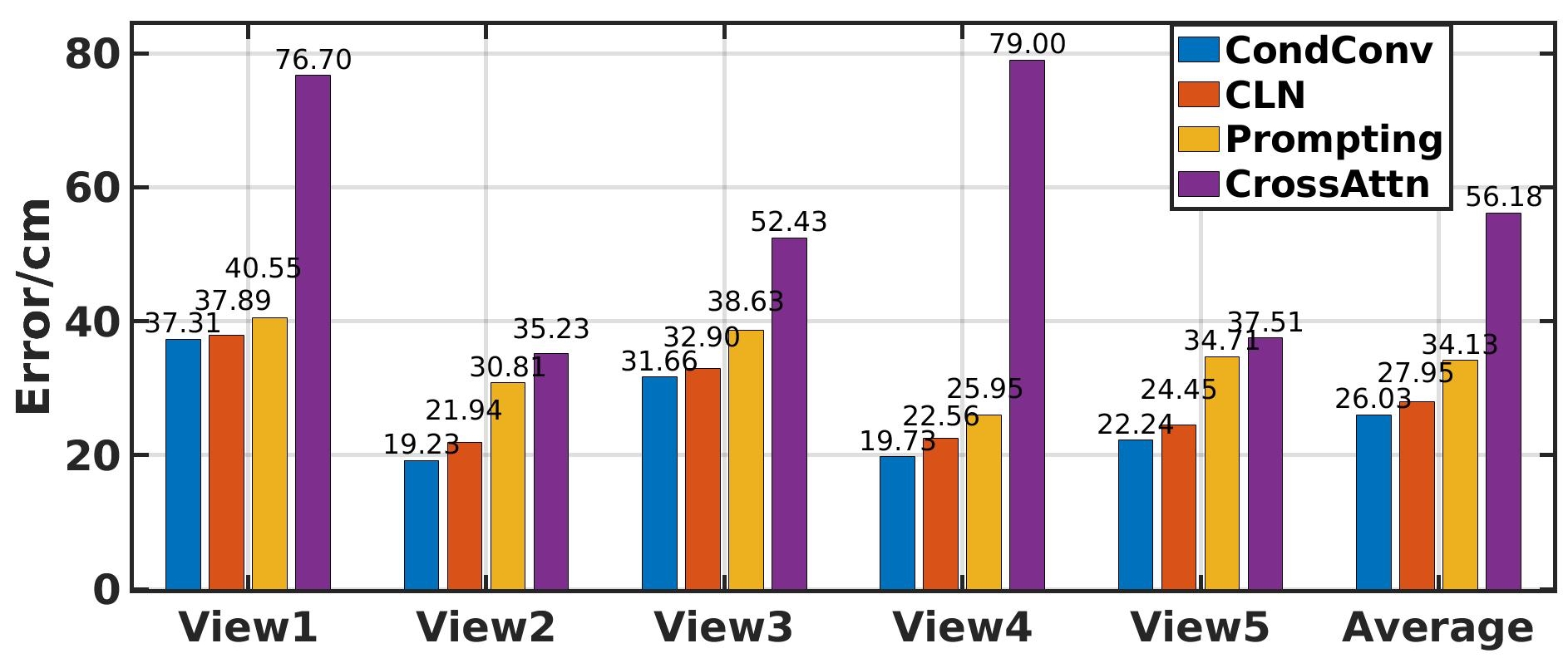}
    \caption{Comparisons of different methods to inject pose information.}
    \vspace{-17pt}
    \label{fig:cond_compare}
\end{figure}

\subsection{Different Methods of Fusing Conditional Information}
We also compare \name's performance against various existing methods of integrating sensor pose information for other applications to showcase the performance benefit of the conditional layers. As the techniques we utilize require transformer encoders, we shift to an alternate model where the ResNets are removed and all the backbones are composed of transformer encoders. We implement and evaluate two networks inspired from adjacent works targeting perspective shift in other contexts, while stressing that these works do not attempt to address the issue of sensor pose shift for \emph{localization}:

\begin{enumerate}
    \item \emph{Cross Attention Pose Fusion} ~\cite{yang2024crossattn}. This work incorporates camera pose into a diffusion model through cross-attention, which grants the model an understanding of how camera location and orientation impacts the video. We utilize this concept to fuse camera pose information through cross attention layers integrated into the transformer backbones.
    \item \emph{Pose Concatenation with Input} ~\cite{zhao2021prompting}. This work concatenates camera pose information into the input to improve monocular depth map prediction on unseen camera viewpoints. We implement their technique by encoding the camera pose as a single token that is concatenated to the sensor data tokens to be fed into the transformer backbones. 
\end{enumerate}

In Figure~\ref{fig:cond_compare}, we observe that integrating pose information through \name (CondConv) and \name-Light (CLN) achieves approximately 20\% improvement over the next best method of concatenating prompts to the input. Furthermore, the Cross Attention method performs poorly with an error of 56 cm. The inferior performance of the other methods demonstrates the importance of carefully selecting the right method for incorporating node pose information.

\subsection{Impact of different training strategies}
\label{subsec:training_strats}

In Table~\ref{table:pretrain}, we analyze the impact of loading pre-trained weights to initialize the network. In Section~\ref{subsec:pretrain}, we introduced a pretraining strategy where we loaded pre-trained weights into the backbone to aid in identifying the car. When training CLN without this strategy, the model performance degrades by 7.1 cm (24\%), highlighting the importance of starting from a good set of weights for CLN. In contrast, CondConv was not adversely impacted with a performance loss of only 0.7 cm. The difference between these two models is likely due to CLN's position within the backbone, where it is closely coupled with the backbone weights and thus highly sensitive to their values. CondConv is placed \emph{after} the backbone layers and is more insulated from the backbone weights.\looseness=-1

\begin{figure}[t]
    \setlength{\abovecaptionskip}{-0.2cm}
    \centering
    \includegraphics[width=0.9\linewidth]{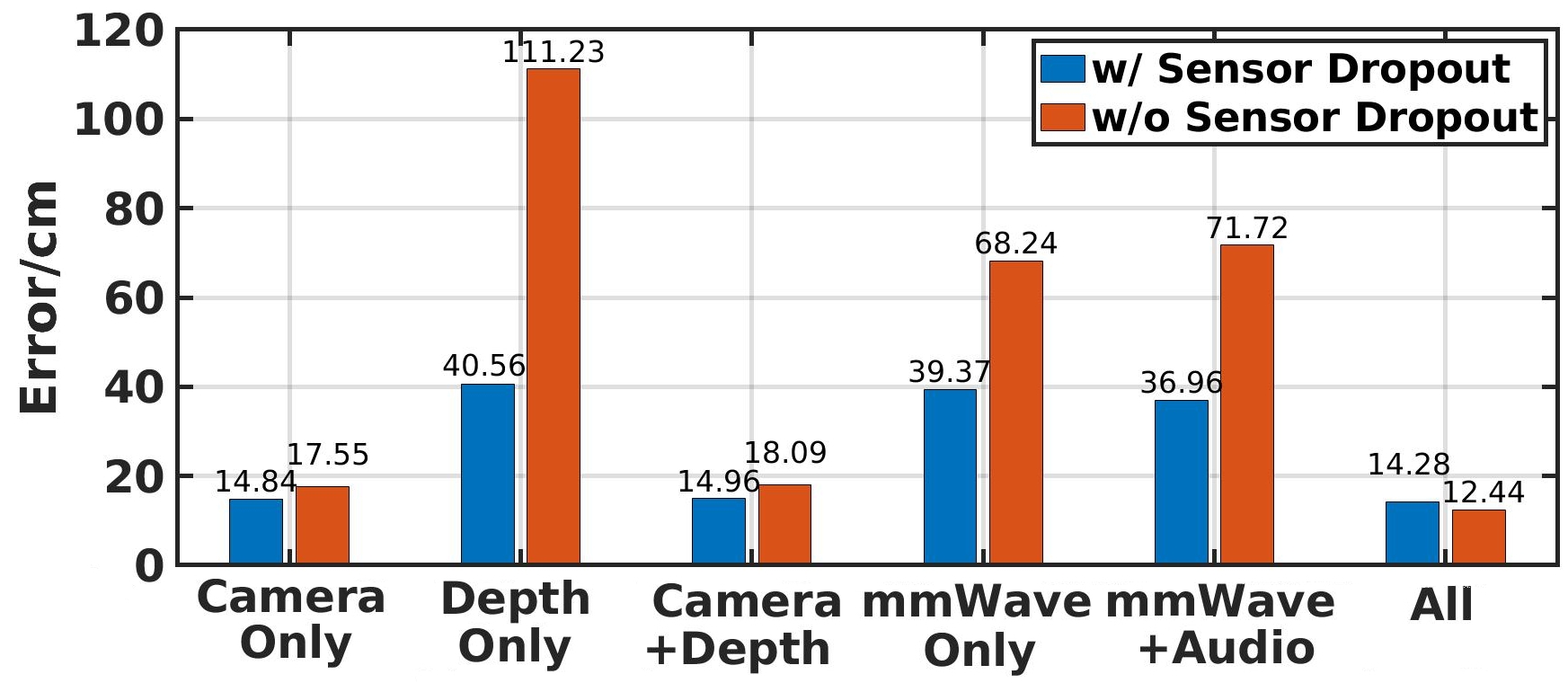}
    \caption{\name performance with and without dropout when only a subset of sensing modalities are available.}
    \vspace{-5pt}
    \label{fig:dropout}
\end{figure}

Figure \ref{fig:dropout} illustrates the impact of the \emph{sensor dropout} training method outlined in Section~\ref{subsec:pretrain}. Without sensor dropout, the model over-relies on the camera modality and largely ignores other modalities, incurring catastrophic error when testing on combinations of modalities without camera (150~\!cm for depth  and 60~\!cm for mmWave). Applying sensor dropout increases the robustness of the model to missing modalities, allowing it to maintain good localization performance despite the loss of the critical camera modality.

\begin{table}[t!]
\begin{tabular}{|c|c|c|}
\hline
                    & From Pretrained Backbones & From Scratch \\ \hline
FlexLoc (ConcConv)  & \textbf{14.28}            & 14.92        \\ \hline
FlexLoc-light (CLN) & \textbf{22.04}            & 29.14        \\ \hline
\end{tabular}
\caption{Effect of Pretraining}
\vspace{-25pt}
\label{table:pretrain}
\end{table}


\begin{table}[b!]
\centering
\vspace{-10pt}
\begin{tabular}{|c|c|cc|}
\hline
                    & Performance                                                 & \multicolumn{2}{c|}{Additional Overhead}                                                                                                     \\ \hline
Models              & Localization Error                                          & \multicolumn{1}{c|}{Parameters}                                                 & MACS (M)                                                     \\ \hline
FlexLoc (CondConv)  & \begin{tabular}[c]{@{}c@{}}14.28\\ (-46.84\%)\end{tabular}  & \multicolumn{1}{c|}{\begin{tabular}[c]{@{}c@{}}53904 \\ (+0.19\%)\end{tabular}} & \begin{tabular}[c]{@{}c@{}}95.46\\ (+0.57\%)\end{tabular}  \\ \hline
FlexLoc-light (CLN) & \begin{tabular}[c]{@{}c@{}}22.04 \\ (-17.96\%)\end{tabular} & \multicolumn{1}{c|}{\begin{tabular}[c]{@{}c@{}}-11788\\ (-0.04\%)\end{tabular}} & \begin{tabular}[c]{@{}c@{}}-54.04\\ (-0.32\%)\end{tabular} \\ \hline
CondConv + CLN      & \begin{tabular}[c]{@{}c@{}}16.11 \\ (-40.02\%)\end{tabular} & \multicolumn{1}{c|}{\begin{tabular}[c]{@{}c@{}}42116 \\ (+0.15\%)\end{tabular}} & \begin{tabular}[c]{@{}c@{}}41.28 \\ (+0.24\%)\end{tabular} \\ \hline
\end{tabular}
\caption{Additional overhead incurred by \name in terms of parameters and computations.}
\label{table:params}
\end{table}


\subsection{Number of Extra Network Parameters}
\label{subsec:params}
Both of the conditional components utilize a set of linear layers to derive their weights from the provided pose information. CondConv utilizes a wider set of linear layers with a total of 52992 parameters, while CLN is more lightweight with only 3328 parameters. However, CLN \textbf{replaces} the BatchNorm2D layers within the ResNet-18 architecture that each have more learnable $\mathbf{\gamma}$ and $\mathbf{\beta}$ parameters (Section~\ref{subsec:cln}), resulting in \emph{fewer model parameters compared to a network without conditional layers}. Table~\ref{table:params} illustrates that CondConv results in 0.19\% more model parameters and a computational overhead of 0.57\% more Multiply-and-Accumulate operations (MACs). In contrast, CLN removes parameters from the network, containing 0.04\% fewer parameters and 0.32\% fewer MACs. In smaller networks where these parameters and operations are more significant, CLN will be a better choice than CondConv despite the worse performance. Interestingly, unifying both CondConv and CLN into the same network resulted in worse performance than CondConv alone. This is likely because CLN and CondConv integrate similar information into the network, causing a saturation of performance, and the increased difficulty of learning with two conditional components hurts performance. Due to the similar overhead and worse performance of the unified model compared to CondConv, we choose to keep CLN and CondConv as separate architectures.

\section{Related Work}
\label{sec:relatedwork}


\subsection{Multimodal Learning for Localization and Tracking} Machine learning algorithms based on handcrafted features and deep neural networks have been applied to indoor localization and tracking applications. For example,
\emph{Samplawski et al.}~\cite{samplawski2023heteroskedastic} used three different cameras to predict the distribution of the object’s location, and designed a \emph{multi-view} late fusion neural network to fuse the prediction of different cameras. Although they demonstrated excellent tracking results, the solution utilizes only RGB cameras and will fail under low-lighting conditions. Therefore, several new multimodal sensing systems have been proposed to leverage the strengths of different sensor modalities to enhance the reliability of indoor localization and tracking \cite{liu2017multiview, zhao2019enhancing,amri2015indoor}. For instance, \cite{liu2017multiview} proposed a method to locate an indoor mobile phone user based on multi-view and multi-modal measurements, but incurs a 1-meter error.

However, aside from \cite{wang2023gdtm}, there exists minimal work in the field of multimodal and multiview localization through localization infrastructures. 
Existing work typically revolves around either unimodal multi-view localization infrastructures \cite{salimibeni2020ble, ngamakeur2023pir}, multimodal single-view localization infrastructures \cite{kandylakis2019fusing, nakamura2011SSL}, or localization through egocentric sensors \cite{caesar2020nuscenes, sun2020scalability}. The few multimodal and multiview works ~\cite{bocus2022operanet, torres2018sleep} do not contain shifting sensor viewpoints or a large diversity of multimodal sensors. Thus, we were unable to perform evaluations on different datasets or find similar architectures that attempted to address the challenge of perspective shift in multimodal multi-view localization.


\subsection{Conditional Neural Networks}

Conditional neural networks incorporate additional auxiliary information during the training of the network to enhance the robustness or accuracy of predictions, demonstrating excellent performance in applications such as Visual Question Answering or Speech Enhancement~\cite{vries2017CBN, zhang2021CLN, strub2018FILM}. For example, De Vries et. al \cite{vries2017CBN} introduces conditional batch normalization to modulate convolutional feature maps of visual processing with a linguistic embedding, which significantly improves the performance of visual question-answering tasks. Zhang et. al \cite{zhang2021CLN} integrates a deep autoencoder with neural noise embedding for speech enhancement through conditional layer normalization, which improves the generalization of the trained model to various noisy speech inputs. Therefore, by training with a small portion of additional weights conditioned on the auxiliary input, the conditional neural networks improve the model accuracy and generalization ability. \name is the only work that applies conditional neural networks to the task of perspective invariant localization.

\section{Limitations, Future Work, and Discussions}

\subsubsection{Sensor Pose Estimation from Sensing Data}

We currently measure the node pose with an OptiTrack motion capture system and provide this ground truth measurement at test time for an new configuration of node viewpoints. Unfortunately, such motion capture systems are not readily available in practical deployment scenarios due to cost and calibration requirements. We can address this lack of external pose information by implementing self-localization techniques using \name's on-board sensor suite to derive its own pose information by employing AprilTag markers~\cite{olson2011AprilTag} or through markerless techniques such as iNERF~\cite{chen2021inerf}. The addition of these techniques will create a fully end-to-end and self-contained localization system that does not depend on any additional infrastructure for perspective invariance.

\subsubsection{Extension to More Comprehensive Environments}
While we achieved exemplary localization results in the presence of sensor perspective shift, we acknowledge that the setting of the dataset (single car driving around an indoor track) is fairly simplistic. Future work can explore training conditional neural networks on datasets containing more complex scenes, from indoor environments with diverse settings and occlusions, to outdoor environments with various weather conditions or landscapes. Furthermore, it will be interesting to evaluate \name's performance in the case of \emph{multi-object localization} by replacing the final transformer encoder with a transformer decoder similar to \emph{DETR}~\cite{carion2020detr}.

\subsubsection{Comparison to Late Fusion Architectures}
While we primarily employed an early-fusion strategy due to superior performance, late fusion can be a superior option in certain scenarios. 
In a distributed computing setting, late fusion has its traditional advantage of flexible composition (sensors make independent predictions, so the number of required sensors is flexible) and minimal communication overhead (only the final prediction results need to be centralized).  
Apart from achieving better performance, we also tried to absorb these advantages of late fusion into the design of \name. First, we used a transformer (encoder) architecture to process the concatenated sensor features. Our system can then handle a dynamic subset of sensor data thanks to the ability of transformers to process arbitrary lengths of features. 
Second, \name leverages a backbone-adapter-fusion architecture. Each node can either offload the raw sensor data to the central server or compute the features locally and then stream the features (compact embeddings of size 256, which reduces the communication overhead). 
We expect feature work to investigate optimal strategies for arranging communications and computations based on the capacity of the nodes and servers.


\section{Conclusions}

In this paper, we propose \name, a multi-view multimodal object localization system that is robust to sensor perspective shifts. \name employs \emph{conditional neural networks} to inject node perspective information into the localization pipeline, which can automatically adapt to unseen sensor perspectives without additional calibration data, achieving zero-shot generalization. Our evaluations on a self-collected multimodal, multi-view indoor tracking dataset show that \name outperforms existing baselines by almost 50\% in localization accuracy while incurring minimal additional overhead.\looseness=-1




\newpage
\section*{ACKNOWLEDGMENT}
The research reported in this paper was sponsored in part by: the IoBT REIGN Collaborative Research Alliance funded by the Army Research Laboratory (ARL) under Cooperative Agreement W911NF-17-2-0196;  the National Science Foundation (NSF) under award \#1822935; and the Air Force Office of Scientific Research (AFOSR) under award FA9550-22-1-0193. The views and conclusions contained in this document are those of the authors and should not be interpreted as representing the official policies, either expressed or implied, of the funding agencies.

\bibliographystyle{IEEEtran}
\bibliography{reference}

\end{document}